\pdfoutput=1

\documentclass[11pt]{article}

\usepackage{acl}

\usepackage{times}
\usepackage{latexsym}

\usepackage[T1]{fontenc}

\usepackage{CJKutf8}
\usepackage[utf8]{inputenc}

\usepackage{microtype}

\usepackage{inconsolata}

\usepackage{soul}
\usepackage{subfigure}
\usepackage{bm}
\usepackage{amsmath}
\usepackage{amssymb}
\usepackage{tablefootnote}
\usepackage{multirow}
\usepackage{booktabs}
\usepackage{arydshln}
\usepackage{tikz}
\usetikzlibrary[arrows.meta, decorations.pathmorphing, backgrounds, positioning, fit, petri, graphs, math, external, calc, shapes, shapes.geometric, decorations.text, shadings]
\usepackage{tikz-dependency}
\usepackage{graphicx}
\usepackage{xcolor}
\usepackage{microtype}
\usepackage{mathtools}
\usepackage{adjustbox}
\usepackage{enumerate}
\usepackage{pgfplots}
\usepackage{fontawesome}
\usepackage{xpinyin}
\usepackage{pifont}

\definecolor{myorange}{HTML}{FFB266} %
\definecolor{mygreen}{HTML}{82CA9D}  %
\definecolor{myred}{HTML}{F08080}    %

\title{Improving Contextual ASR via Multi-grained Fusion \\ with Large Language Models}

\author{
    Shilin Zhou$^{1}$,
    Zhenghua Li$^{1}$\Thanks{$~$ Corresponding author}\\
    $^1$School of Computer Science and Technology, \\
    Soochow University, Suzhou, China \\
    \texttt{slzhou.cs@outlook.com; zhli13@suda.edu.cn} \\
}

\pgfplotsset{compat=1.17}
\begin{document}
\begin{CJK}{UTF8}{gkai}
\maketitle
\begin{abstract}
    While end-to-end Automatic Speech Recognition (ASR) models have shown impressive performance in transcribing general speech, they often struggle to accurately recognize contextually relevant keywords,  such as proper nouns or user-specific entities. 
    Previous approaches have explored leveraging keyword dictionaries in the textual modality to improve keyword recognition, either through token-level fusion that guides token-by-token generation or phrase-level fusion that enables direct copying of keyword phrases. 
    However, these methods operate at different granularities and have their own limitations. 
    In this paper, we propose a novel multi-grained fusion approach that jointly leverages the strengths of both token-level and phrase-level fusion with Large Language Models (LLMs).  
    Our approach incorporates a late-fusion strategy that elegantly combines ASR's acoustic information with LLM's rich contextual knowledge, balancing fine-grained token precision with holistic phrase-level understanding. 
    Experiments on Chinese and English datasets demonstrate that our approach achieves state-of-the-art performance on keyword-related metrics while preserving high accuracy on non-keyword text. 
    Ablation studies further confirm that the token-level and phrase-level components both contribute significantly to the performance gains, complementing each other in our joint multi-grained framework.
    The code and models will be publicly available at \url{https://github.com/}.
\end{abstract}

\section{Introduction}
Current end-to-end Automatic Speech Recognition (ASR) models, such as Whisper \cite{radford-2022-whisper, gandhi-2023-distil}, have demonstrated impressive performance in transcribing common words. 
However, these models often struggle to accurately transcribe contextually relevant keywords \cite{alon-2019-hard-example, yang-2024-mala, zhou-2023-copyne}. 
Such keywords may include proper nouns or user-specific entities, such as contact names from a phone's address book. 
These keywords often convey important semantics, which are essential for understanding a sentence's meaning and performing related tasks accurately.
In many scenarios, contextual keywords frequently occur within predefined contexts, such as contact names stored in a phone or previously searched terms. This contextual information is typically readily accessible. 
To improve the recognition of these keywords during transcription, prior research has explored the use of contextual keyword dictionaries \cite{pundak-2018-clas, alon-2019-hard-example, lakomkin-2024-end}.

Large language models (LLMs) \cite{openai-2023-gpt4,  touvron-2023-llama, yang-2023-baichuan, abdin-2024-phi, yang-2024-qwen2-5, kwon-2023-efficient} have recently demonstrated exceptional capabilities in contextual modeling and reasoning across diverse tasks. 
These strengths align closely with the demands of contextual ASR, where the accurate recognition of user-relevant keywords heavily depends on contextual information. 
Consequently, there has been growing interest in leveraging LLMs to help ASR model better recognize keywords \cite{sun-2024-contextual, lakomkin-2024-end, yang-2024-mala}.
Typically, contextual keyword dictionaries in the textual modality are provided to LLMs in the form of a prompt $C$. Based on this, \citet{sun-2024-contextual} proposed a two-pass approach. In the first pass, the ASR model generates multiple transcriptions. These candidates are then fed into the LLM, which computes second-pass scores conditioned on $C$. The LLM calculates the likelihood of each token sequentially during decoding and aggregates these token-level scores to produce an overall score for each candidate. 
The candidate with the highest score is selected as the final result.
In addition, \citet{lakomkin-2024-end} and \citet{yang-2024-mala} proposed end-to-end based methods. 
Beyond providing the contextual prompt $C$, these methods directly feed audio input into the LLM via adapters and train the model to predict transcription text. 
Similar to the two-pass approach, the decoding process in end-to-end methods operates at the token level, where the LLM generates transcription text in a token-by-token, autoregressive manner.
Both approaches share the characteristic of relying on token-level operations during decoding.
For this reason, we categorize them as token-level fusion approaches in this paper.

However, keywords are often multi-token phrases, and token-level fusion approaches lack a holistic understanding of the entire keyword phrase. 
This limitation can result in incomplete generation of keywords. 
To address this issue, \citet{zhou-2023-copyne} proposed a phrase-level fusion approach, introducing a phrase-level copy loss to guide the ASR model in selecting the correct keyword phrase. 
This approach allows the model to directly copy all tokens of a keyword phrase from a predefined dictionary, significantly improving the overall recall of keywords.
However, the effectiveness of the phrase-level approach depends on the accurate selection of keywords from the dictionary. They relied exclusively on speech information from the ASR model for keyword selection, which may result in incorrect choices. 
This, in turn, may adversely affect the recognition accuracy of non-keyword text.

Both token-level and phrase-level fusion approaches have demonstrated their effectiveness, yet they operate at different granularities.
Token-level fusion helps guide the model to generate individual tokens more accurately by leveraging the contextual understanding of the LLM, while phrase-level fusion ensures the completeness and coherence of keyword phrases through direct copying mechanisms.
However, there has not been a unified framework that effectively combines these different granularities to leverage their respective strengths. 
A joint approach that can simultaneously benefit from token-level precision and phrase-level coherence could potentially lead to more robust and accurate keyword recognition.

\begin{figure*}[tb]
    \centering
    \scalebox{0.8}{
    \includegraphics[width=1.0\textwidth]{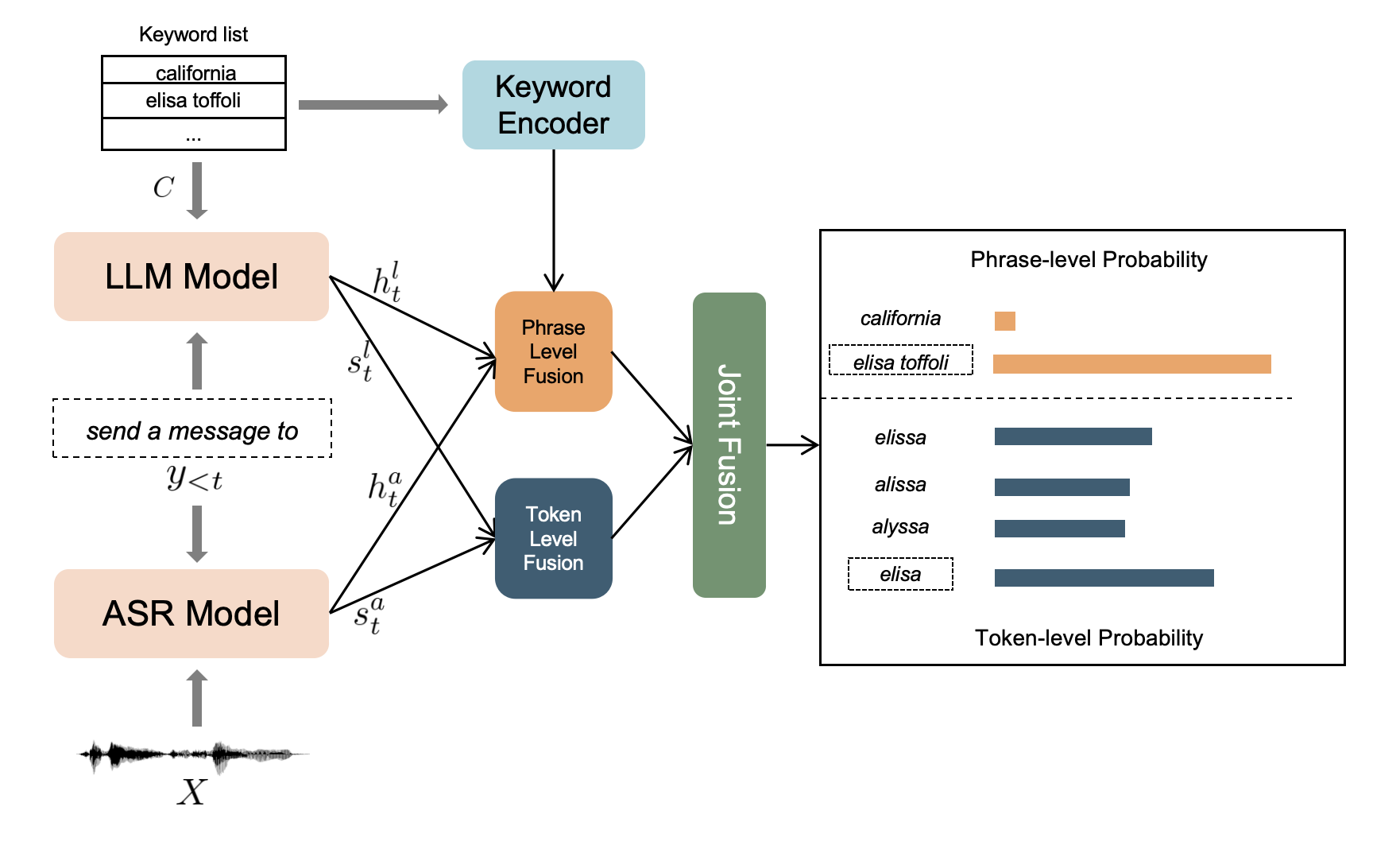}
    }
    \caption{
        An example of transcribing the speech "send a message to elisa toffoli". Here, $X$ is the speech input. $C$ is the contextual prompt. $y_{<t}$ is the previously generated text. $s_{t}^{l}, h_{t}^{l}$ and $s_{t}^{a}, h_{t}^{a}$ are the logit score and hidden state of the LLM and ASR model, respectively.
        }
    \label{fig:framework}
\end{figure*}

In this paper, we propose a multi-grained fusion approach that jointly fuses the speech information from ASR and the contextual information from LLMs at both token-level and phrase-level to improve ASR keyword recognition.
Figure \ref{fig:framework} illustrates the framework of our joint approach.
The overall process is as follows: the keyword list in text modality is first provided to the LLM in the form of a prompt.
During transcription, the knowledge of ASR and LLM is fused in a late-fusion manner to improve keyword recognition.
Specifically, in our token-level fusion, we fuse the token-level logit scores from the ASR and LLM to guide the model in generating tokens. 
In phrase-level fusion, we further fuse the hidden representations of the ASR and LLM to enable the model to select the correct keyword phrase from the dictionary.
Finally, through the multi-grained mechanism, our approach harmoniously integrates the fine-grained precision of token-level fusion with the holistic contextual awareness of phrase-level fusion. 
This integration yields transcriptions that are both more accurate for individual tokens and semantically coherent for complete keyword phrases.

We conduct experiments on recently released Chinese and English datasets \cite{chen-2022-aishell, zhou-etal-2024-chinese, wang-2024-slidespeech}.
Experiments demonstrate that our proposed approach achieves performance comparable to state-of-the-art methods, particularly on keyword-related metrics, while maintaining the accuracy on non-keyword text.
Ablation studies show that both the token-level and phrase-level components play crucial roles in the joint multi-grained fusion mechanism.
The contributions can be summarized as follows:
\begin{itemize}
    \item We propose a multi-grained fusion approach that seamlessly integrates speech and text modalities at both token-level and phrase-level to improve the recognition of contextually relevant keywords. 
    \item Our framework harnesses the powerful contextual reasoning capabilities of LLMs to effectively combine acoustic evidence with linguistic knowledge, resulting in more accurate and contextually appropriate transcriptions.
    \item We evaluate our joint multi-grained fusion approach on Chinese and English datasets. 
    The results show that our approach achieves performance on par with previous approaches and even surpasses them on keyword-related metrics. 
    The code and models will be released on GitHub.
\end{itemize}

\section{Related Work}
\subsection{Contextual Biasing in ASR}
Using keyword dictionaries in textual modality to help ASR models for better transcription has been widely studied. Depending on the granularity of the biasing approach, these approaches can be categorized into token-level and phrase-level approaches.

\subsubsection{Token-level.}
\citet{pundak-2018-clas} introduced a method that encodes each keyword using a dedicated encoder and applies attention to produce a list-level representation, which is fed into the ASR model to bias token-by-token generation. Later works improved robustness via training noise \cite{alon-2019-hard-example}, enhanced keyword encoders \cite{char-level-concder-2023}, or filtered irrelevant keywords \cite{jayanthi-2023-retrieve}.

Beyond encoder-based fusion, TCPGen \cite{sun-2022-tree, sun-2023-can} avoids context encoding and directly manipulates the output distribution by interpolating ASR predictions with a pointer-based bias constrained by a prefix tree. This design enables use with frozen ASR models, training only the fusion module on task-specific data using decoder hidden states.

However, token-level methods often treat keywords as independent tokens, lacking holistic phrase-level understanding. This can result in incomplete keyword predictions and lower recall.

\subsubsection{Phrase-level.}
Phrase-level approaches aim to generate complete keyword phrases, ensuring their integrity compared to token-level methods.
CopyNE \cite{zhou-2023-copyne} introduced a copy mechanism into ASR, allowing the model to copy keywords directly from a predefined list.
Similarly, \citet{sudo-2024-contextualized} extended the token vocabulary by merging it with the keyword list, enabling direct phrase prediction during decoding.

While phrase-level approaches improve keyword accuracy, they may degrade non-keyword recognition.
In contrast, our joint biasing method combines token-level and phrase-level fusion, improving keyword accuracy without sacrificing non-keyword performance.

\subsection{Using LLMs in ASR}
Language models (LMs) encode rich linguistic knowledge through large-scale text pretraining. Traditional n-gram LMs have long been used in ASR to improve fluency \cite{yao-2021-wenet, chorowski-2016-shallow, sriram-2017-cold}.
With the rise of large language models (LLMs), their strong contextual and reasoning capabilities have led to growing interest in using LLMs to enhance ASR, especially for keyword recognition. Existing methods can be broadly divided into end-to-end and two-pass approaches.

\subsubsection{End-to-end.}
End-to-end approaches \cite{yang-2024-mala, lakomkin-2024-end, bai-2024-seedasr} train LLMs to generate transcriptions from speech and auxiliary text in a single stage.
For example, \citet{yang-2024-mala} proposed a contextual ASR model where the keyword list and speech features (extracted via a speech encoder) are concatenated and fed into an LLM for one-pass transcription.

\subsubsection{Two-pass.}
Two-pass approaches first generate candidate transcriptions, then refine them using LLMs.
\citet{sun-2024-contextual} applied LLM-based rescoring to select the best hypothesis.
\citet{chens-2024-its} and \citet{hu-2024-large}
further incorporated both audio and first-pass candidates as input to the LLM, enabling joint reasoning over speech and text.

In this work, we propose a novel end-to-end multi-grained fusion approach that integrates LLMs into the decoding process to guide keyword recognition.
Unlike two-pass methods, our model performs keyword-aware decoding in a single stage, allowing direct LLM intervention during transcription generation.

\section{Proposed Approach}
Our multi-grained fusion approach leverages the complementary strengths of token-level and phrase-level fusion via a late-fusion framework with an LLM.
We first inject contextual keyword knowledge into the LLM and then use its output to bias the ASR model toward accurate keyword transcription.
The proposed joint multi-grained fusion framework comprises four components: knowledge injection, token-level fusion, phrase-level fusion, and their integration.

\subsection{Knowledge Injection}
Given a keyword list $K = (k_1, k_2, ..., k_N)$, we construct a contextual prompt $C$ by concatenating the keywords and instructing the LLM to focus on relevant ones during transcription. The prompt is formulated as:
\emph{Transcribe the speech into text. The following keywords are likely to appear. Use relevant keywords to improve transcription accuracy and ignore irrelevant ones. The keywords are {$k_1$, $k_2$, ..., $k_N$}. The text corresponding to the speech is:}

Although prior work also incorporates keyword prompts into the LLM \cite{li-2023-prompting, sun-2024-contextual, lakomkin-2024-end, yang-2024-mala}, these methods typically require additional inputs to the LLM, such as candidate transcriptions in two-pass systems or audio representations in end-to-end approaches.
In contrast, our approach uses only the keyword prompt, enabling efficient one-pass decoding without LLM fine-tuning or audio encoding.

\subsection{Token-level Fusion}
Token-level fusion integrates the semantic information from the LLM and the acoustic information from the ASR model during the token-by-token decoding process.
Specifically, at decoding step $t$, the LLM produces the logit score $\bm{s}_{t}^{l}$ 
for all tokens in the token vocabulary $\mathcal{V}$ based on the contextual prompt $C$ and the previously generated text $y_{<t}$.
Similarly, the ASR model produces the logit score $\bm{s}_{t}^{a}$ for the next token based on $y_{<t}$ and the speech input $X$.
The $\bm{s}_{t}^{l}$ and $\bm{s}_{t}^{a}$ model the probabilities for the next token from semantic and acoustic perspectives, respectively. 
\begin{equation}\label{eq:llm}
    \bm{h}_{t}^{l}, \bm{s}_{t}^{l}  = \text{LLM}(y_{<t}, C)
\end{equation}
\begin{equation}\label{eq:asr}
    \bm{h}_{t}^{a}, \bm{s}_{t}^{a}  = \text{ASR}(y_{<t}, X)
\end{equation}
Here, LLM could be a pre-trained decoder-only transformer model, such as GPT-3 \cite{mann-2020-language}, Llama \cite{touvron-2023-llama}, or Qwen \cite{yang-2024-qwen2-5}.
The ASR model can be an encoder-decoder transformer model. In this work, we use the pre-trained Whisper \cite{radford-2022-whisper} as the ASR model.
$\bm{h}_{t}^{l}$ and $\bm{h}_{t}^{a}$ are the hidden states of the LLM and ASR model.

We combine $\bm{s}_{t}^{l}$ and $\bm{s}_{t}^{a}$ to compute the final probability of generating next token.
However, since $\bm{s}_{t}^{l}$ and $\bm{s}_{t}^{a}$ originate from different modalities, and the computation of the two scores relies on different information, they cannot be directly compared.
For example, the computation of $\bm{s}_{t}^{l}$ depends solely on $y_{<t}$ and $C$. 
At the beginning of generation, $y_{<t}$ contains limited information, causing the LLM to predict within a broad semantic space.
Excessive reliance on $\bm{s}_{t}^{l}$ may lead to the model deviating from the true transcription corresponding to the speech input.
On the contrary, $\bm{s}_{t}^{a}$ is based on the speech input $X$ and $y_{<t}$, which may align better with the speech contents.
Therefore, inspired by \citet{chens-2024-its}, we employ a score fusion strategy based on uncertainty to balance the predictions of the LLM and ASR. 
The final token-level fusion formulas are as follows:
\begin{equation}
    \bm{s}_{t} = \bm{s}_{t}^{a} + \text{sigmoid}(\bm{u}_{t}^{a}) \cdot \bm{s}_{t}^{l}
\end{equation}
\begin{equation}\label{eq:uncertainty}
    \bm{u}_{t}^{a} = -\bm{p}_{t}^{a} \cdot \log (\bm{p}_{t}^{a})
\end{equation}
\begin{equation}\label{eq:tok}
    p_{tok}(y_{t} | y_{<t}, C, X) = \left[\text{softmax}(\bm{s}_{t})\right]_{y_{t}}
\end{equation}
Here, $\bm{u}_{t}^{a}$ represents the uncertainty of the ASR model. $\bm{p}_{t}^{a}$ is the probability of the ASR model obtained from applying softmax to $\bm{s}_{t}^{a}$.
It means that when the ASR model has high uncertainty, for example, if a keyword is difficult to distinguish based on speech information alone, more weight is given to the LLM's prediction. Conversely, at the beginning of the generation, when the ASR model is more confident, greater weight is assigned to the ASR's prediction.
$\bm{s}_{t}$ and $p_{tok}(y_{t} | y_{<t}, C, X)$ represent the final fused token-level score and probability, respectively.

\subsection{Phrase-level Fusion}
Unlike the token-level fusion, which applies biasing to each individual token in the vocabulary, the phrase-level fusion treats each keyword as a whole unit,
and biases the model to focus on entire key phrases, enabling it to transcribe target keywords as cohesive units.

Specifically, we design a  fusion module to integrate the phrase-level representation of each keyword, the contextual semantic information from the LLM, and the speech information from the ASR to determine the phrase-level probability for each keyword.
At decoding step $t$, the module estimates the probability of generating a complete phrase from the keyword list.

First, we obtain the phrase-level representation $\bm{r}_i$ for each keyword $k_i$ in the keyword list $K$ using a keyword encoder. 
Following CopyNE \cite{zhou-2023-copyne}, we use a 3-layer LSTM as the keyword encoder to encode each keyword.
We take the hidden state of the last token from the last layer as the phrase-level representation $\bm{r}_i$ for each keyword $k_i$.
\begin{equation}
    \bm{r}_i = \text{LSTM}(k_i)
\end{equation}

Then, for contextual semantic information 
and acoustic information, we use the hidden states $\bm{h}_{t}^{l}$ and $\bm{h}_{t}^{a}$ from the LLM and ASR model, as shown in equation \ref{eq:llm} and \ref{eq:asr}.
With $\bm{r}_i$, $\bm{h}_{t}^{l}$, and $\bm{h}_{t}^{a}$ obtained, we compute the phrase-level probability $p_{phr}(k_i | y_{<t}, C, X)$ for each keyword $k_i$ using a dot-product attention mechanism.

First, $\bm{h}_{t}^{l}$ and $\bm{h}_{t}^{a}$ are concatenated and passed through a linear layer to obtain the attention query $\bm{q}_{t}$.
Then, we calculate the attention score between the query $\bm{q}_{t}$ and each phrase-level representation $\bm{r}_i$.

Finally, a softmax function is applied to obtain the phrase-level probability $p_{phr}(k_i | y_{<t}, C, X)$.
\begin{equation}
    \bm{q}_{t} = \text{Linear}([\bm{h}_{t}^{l}; \bm{h}_{t}^{a}])
\end{equation}
\begin{equation}\label{eq:phrase}
    p_{phr}(k_i | y_{<t}, C, X) = \frac{\exp(\bm{q}_{t} \cdot \bm{r}_i)}{\sum_{j=1}^{N} \exp(\bm{q}_{t} \cdot \bm{r}_j)}
\end{equation}

\subsection{Joint Multi-grained Fusion}
The token-level probability $p_{tok}(y_{t} | y_{<t}, C, X)$ guides the model on which token to generate next, while the phrase-level probability $p_{phr}(k_i | y_{<t}, C, X)$ informs the model to pick an entire keyword phrase. 
Different choices between generating individual tokens and entire keywords can result in varying final outcomes.
The two probabilities are computed based on different information from different granularity levels.
So it is important to jointly fuse the two probabilities to make the best transcription.
However, it is non-trivial to directly combine the two probabilities, as they are computed based on different information and have different scales.
To unify these two probabilities of different granularities into a comparable scale where they can effectively complement each other, we further propose a simple yet effective multi-grained fusion approach.

Specifically, we add a fake keyword $k_0$ to the keyword list $K$. 
When the model should generate tokens rather than select a whole keyword, we train it to predict $k_0$.
So the equation \eqref{eq:phrase} becomes:
\begin{equation}\label{eq:phr}
    p_{phr}(k_i | y_{<t}, C, X) = \frac{\exp(\bm{q}_{t} \cdot \bm{r}_i)}{\sum_{j=0}^{N} \exp(\bm{q}_{t} \cdot \bm{r}_j)}
\end{equation}
$p_{phr}(k_0 | y_{<t}, C, X)$, the phrase-level probability of the fake keyword $k_0$, means the probability that the model should generate tokens in the next step rather than selecting a whole keyword.
Accordingly, it can be treated as the prior probability of generating tokens.
Finally, the two sorts of probability, i.e., $p_{tok}(y_{t} | y_{<t}, C, X)$ and $p_{phr}(k_i | y_{<t}, C, X)$, can be jointly normalized to the same scale.
\begin{equation}\label{equ:q}
    p_{joi}(z_i| y_{<t}) = \begin{cases} p_{phr}(k_0|y_{<t})p_{tok}(z_i|y_{<t}),  z_i \in \mathcal{V} \\
    p_{phr}(z_i|y_{<t}), ~~~~~z_i \in K, z_i \neq k_0
                                    \end{cases}
\end{equation}
Here, $\mathcal{V}$ is the token vocabulary. $z_i$ is a token from $\mathcal{V}$ or a keyword from $K$.
For abbreviation, we omit the textual condition $C$ and speech condition $X$ in the equation.
After normalization, tokens in the vocabulary and keywords in the keyword list form a unified multi-grained space $\mathcal{Z} = \mathcal{V} \cup K$.
The probabilities from token-level and phrase-level fusion are jointly normalized in $p_{joi}(z_i|y_{<t})$.
We finally generate the final transcription by doing beam search over $p_{joi}(z_i|y_{<t})$.

\section{Experiments}
\subsection{Experimental Setup}
\subsubsection{Datasets.} 
We conduct experiments on Chinese and English datasets.
For Chinese, we use the Aishell dataset \cite{bu-2017-aishell,chen-2022-aishell} and the RWCS-NER dataset \cite{zhou-etal-2024-chinese}.
The Aishell dataset contains 150 hours of clean speeches recorded in quiet environments. 
RWCS-NER is a recently released test set proposed to evaluate the performance of Chinese Spoken NER in real-world scenarios. 
It covers two domains: open-domain daily conversations (DC) and task-oriented intelligent cockpit instructions (ICI). 
Both the Aishell and RWCS-NER datasets are annotated with transcriptions and the named entities.
We utilize the entities as the corresponding keyword list.

For English, we use the Slidespeech dataset \cite{wang-2024-slidespeech}, a large-scale corpus enriched with slide information.
Each speech sample is paired with its corresponding transcription as well as a keyword list extracted from the associated slide.
We use the 473-hour L95 subset of Slidespeech as the training set.

\subsubsection{Backbone Models.}
Our multi-grained model consists of two main components: an ASR model and a language model (LLM).
For the ASR component, we adopt the Whisper model\footnote{https://huggingface.co/openai/whisper-small} \cite{radford-2022-whisper}, a widely used Transformer-based autoregressive model trained on large-scale multilingual ASR data. It has demonstrated strong performance across many benchmarks.
For the LLM component, we use Qwen2-1.5B \cite{yang-2024-qwen2} for Chinese experiments and Phi-3.5-mini \cite{abdin-2024-phi} for English experiments. Although a single multilingual LLM could, in principle, handle both languages, we opt for language-specific models for two main reasons. First, this choice reflects community preferences, as Qwen2 and Phi-3.5 are widely used and well-optimized for Chinese and English tasks, respectively. Second, using smaller models tailored to each language allows for more efficient inference. It is worth noting that our method is language-agnostic and can seamlessly work with any multilingual LLM if needed.

\begin{table*}[!tb]
        \centering
        \begin{small}
        \setlength{\tabcolsep}{2pt}
        \begin{tabular}{l c cccc cccc cccc}
            \toprule
            \multirow{2}{*}{Model} & \multirow{1}{*}{Using Textual} & 
            \multicolumn{4}{c}{Aishell $_{\mathrm{\text{Entity Char Ratio}} = 9.58\%}$} & \multicolumn{4}{c}{DC $_{\mathrm{\text{Entity Char Ratio}} = 8.34\%}$} & \multicolumn{4}{c}{ICI $_{\mathrm{\text{Entity Char Ratio}} = 18.86\%}$} \\
            \cmidrule(lr){3-6} \cmidrule(lr){7-10} \cmidrule(lr){11-14}
            & Keywords & CER$\downarrow$ & B-CER$\downarrow$ & U-CER$\downarrow$ & R$\uparrow$ & CER$\downarrow$ & B-CER$\downarrow$ & U-CER$\downarrow$ & R$\uparrow$ & CER$\downarrow$ & B-CER$\downarrow$ & U-CER$\downarrow$ & R$\uparrow$ \\
            \midrule
            Whisper & \ding{55} & 5.2 & 10.4 & 4.7 & 80.6 & 12.8 & 22.9 & 11.7 & 71.1 & 11.5 & 30.7 & 6.9 & 40.8 \\
            CopyNE & \ding{51} & 4.6 & 3.4 & 4.7 & 94.4 & 12.2 & 14.9 & 11.8 & 82.0 & 8.9 & 16.8 & 7.0 & 70.0 \\
            Ours & \ding{51} & \bf3.7 & \bf2.2 & \bf3.8 & \bf96.4 & \bf11.0 & \bf11.4 & \bf10.8 & \bf86.6 & \bf7.7 & \bf10.9 & \bf6.8 & \bf79.8 \\
            \hdashline
            Ours w/o P & \ding{51} & 4.3 & 7.0 & 4.0 & 86.9 & 11.6 & 17.9 & \bf10.8 & 77.3 & 10.2 & 20.1 & 7.6 & 61.8 \\
            Ours w/o T & \ding{51} & 4.5 & 2.7 & 4.7 & 95.5 & 12.1 & 13.2 & 11.8 & 84.3 & 8.6 & 14.7 & 7.0 & 73.1 \\

            \bottomrule
        \end{tabular}
        \end{small}
        \caption{Results on Aishell, DC, and ICI test sets. Entity Char Ratio is the ratio of the entity characters in the dataset. w/o P and w/o T indicate the exclusion of the phrase-level and token-level modules, respectively.} 
        \label{table:res_zh}
\end{table*}
\begin{table}[!tb]
    \begin{small}
    \centering
    \setlength{\tabcolsep}{1pt}
    \begin{tabular}{l c cccc}
        \toprule
        \multirow{2}{*}{Model} & \multirow{1}{*}{Using Textual} &
        \multicolumn{4}{c}{Slidespeech $_{\mathrm{\text{Keyword Ratio}} = 6.84\%}$} \\
        \cmidrule(lr){3-6}
        & Keywords & WER$\downarrow$ & B-WER$\downarrow$ & U-WER$\downarrow$ & R$\uparrow$ \\
        \midrule
        Whisper$^{\dagger}$ & \ding{55} & 9.45 & 5.84 & 9.71 & 94.32 \\
        Whisper & \ding{55} & 9.28 & 8.12 & 9.37 & 92.20 \\
        CPP & \ding{51} & 12.38 & 9.32 & 12.60 & 90.86 \\
        LCB-net & \ding{51} & 12.02 & 9.03 & 12.24 & 91.12 \\
        CopyNE & \ding{51} & 9.27 & 6.88 & 9.45 & 93.42 \\
        MaLa-ASR & \ding{51} & 9.14 & 5.47 & 9.42 & 94.87 \\
        Ours & \ding{51} & 9.14 & \bf5.36 & 9.42 & \bf95.18 \\
        \bottomrule
    \end{tabular}
    \caption{Results on the Slidespeech test set. $^{\dagger}$ indicates the model is the original model without fine-tuning.} 
    \label{table:res_eng}
\end{small}
\end{table}

\subsection{Training}
\subsubsection{Loss Function.}
The loss function $\mathcal{L}$ of our model is composed of two parts: the token-level loss $\mathcal{L}_{tok}$ and the phrase-level loss $\mathcal{L}_{phr}$.
First, given the ground-truth transcription $Y={y}_{1}, {y}_{2}, \ldots, {y}_{T}$ of the speech, 
$\mathcal{L}_{tok}$ is the negative log-likelihood loss to generate the transcription $Y$.
\begin{equation}
    \mathcal{L}_{tok} = -\sum_{t=1}^{T} p_{tok}({y}_{t} | {y}_{<t}, C, X)
\end{equation}
where $p_{tok}({y}_{t} | {y}_{<t}, C, X)$ is the token-level probability to generate the token ${y}_{t}$ computed by equation \eqref{eq:tok}.

Second, the phrase-level loss $\mathcal{L}_{phr}$ is used to teach the model to choose the correct keywords from the keyword list $K$.
Given $K$, we apply the maximum matching algorithm to find all the phrases in the transcription $Y$ that also listed in the keyword list $K$. 
Through this process, we can obtain a sequence $P={p}_{1}, {p}_{2}, \ldots, {p}_{T}$. $p_{i} = k_{j}$ if the $j$-th keyword is matched at time step $i$. Otherwise, $p_{i} = k_{0}$, which is a fake keyword indicating no keyword is matched.
So the phrase-level loss $\mathcal{L}_{phr}$ is the loss to generate the phrase-level sequence $P$.
\begin{equation}
    \mathcal{L}_{phr} = -\sum_{t=1}^{T} p_{phr}({p}_{t} | {p}_{<t}, C, X)
\end{equation}
where $p_{phr}({p}_{t} | {p}_{<t}, C, X)$ is the phrase-level probability to generate the phrase ${p}_{t}$ computed by equation \eqref{eq:phr}.

Finally, the total multi-grained loss $\mathcal{L}$ is the sum of $\mathcal{L}_{tok}$ and $\mathcal{L}_{phr}$.
\begin{equation}\label{eq:loss}
    \mathcal{L} = \mathcal{L}_{tok} + \mathcal{L}_{phr}
\end{equation}

\subsubsection{Training Details.}
For Chinese experiments, we train on the Aishell training set and evaluate on the test sets of Aishell, DC, and ICI to assess generalization to out-of-domain data. For English, training is conducted on the L95 subset of SlideSpeech, with evaluation on its test set.
During training, the LLM is frozen. To enable token-level fusion, we replace the ASR model’s tokenizer with that of the LLM \cite{chens-2024-its}. Only the ASR decoder, keyword encoder, and the phrase-level linear layer are updated, totaling ~300M parameters.
We use the AdamW optimizer \cite{loshchilov-2017-fixing} with a learning rate of 1e-4 and batch size of 60. The model is trained for 40 epochs, with 4 epochs of warm-up. Training takes ~40 hours on three A100 40G GPUs. We report the best-performing checkpoint on the validation set.

\subsection{Evaluation}
\subsubsection{Baselines.}
On the Chinese dataset, we first fine-tune the Whisper model \cite{radford-2022-whisper} on the Aishell training set as the initial baseline.
Additionally, the Whisper baseline also denotes as the our joint model with both token-level and phrase-level fusion removed.
CopyNE \cite{zhou-2023-copyne} is the second baseline.
It is a recently proposed state-of-the-art phrase-level fusion model which also utilizes the keyword dictionary to improve the accuracy of keywords.

For English, we additionally compare with CPP \cite{Huang-cpp-2023,wang-2024-slidespeech}, LCB-net \cite{Yu-LCB-2024}, and MaLa-ASR \cite{yang-2024-mala}. CPP is the baseline from the SlideSpeech benchmark. LCB-net introduces a long-context biasing mechanism designed for video transcription. MaLa-ASR is an end-to-end state-of-the-art token-level fusion model with both keyword dictionaries and LLMs.
All models are evaluated with the same keyword list for fairness. Following MaLa-ASR, we set the list size to 50 in main experiments and vary it in ablation studies.

\subsubsection{Metrics.}
Following \citet{zhou-2023-copyne} and \citet{yang-2024-mala}, we report character error rate (CER) for Chinese and word error rate (WER) for English.
To assess keyword performance, we additionally report biased CER/WER (B-CER, B-WER), which measure errors on keyword spans, and keyword recall (Recall), the percentage of keywords fully and correctly recognized.

We also include unbiased CER/WER (U-CER, U-WER), computed on non-keyword segments, to evaluate model impact on general transcription.
To ensure consistency, all references and predictions are normalized using Whisper’s official text normalizer\footnote{github.com/openai/whisper/tree/main/whisper/normalizers} before metric computation.

\subsection{Results}
\begin{figure*}[!tb]
    \centering
    \subfigure[B-CER]{
    \label{fig:effect_keyword_list_bcer}
    \scalebox{0.5}{
    \begin{minipage}{0.45\textwidth}
    \centering
    \begin{tikzpicture}
        \centering
        \begin{axis}[
            xmin=0,xmax=1000,
            ymin=1.5,ymax=9,
            xtick={0, 50, 100, 200, 500, 1000},
            x tick label style={rotate=45,anchor=east},
            ytick={2, 3, 4, 5, 6, 7, 8, 9},
            ymajorgrids=true,
            major grid style={line width=.2pt,draw=gray!50,densely dashed},
            minor grid style={line width=.1pt,line length=.02pt,draw=gray!30},
            legend style={at={(0.75,0.5)}, anchor=south},
            ]
            \addplot[color=myorange, line width=1.5pt, mark=diamond, smooth] coordinates {
            (0, 1.89)
            (50, 2.16)
            (100, 2.50)
            (200, 2.92)
            (500, 3.28)
            (1000, 3.53)
            };
            \addlegendentry{Joint}
            \addplot[color=mygreen, line width=1.5pt, mark=triangle, smooth] coordinates {
            (0, 2.69)
            (50, 2.69)
            (100, 2.90)
            (200, 3.27)
            (500, 3.62)
            (1000, 4.16)
            };
            \addlegendentry{Phrase-level}
            \addplot[color=myred, line width=1.5pt, mark=square, smooth] coordinates {
            (0, 4.08)
            (50, 7.04)
            (100, 7.60)
            (200, 7.85)
            (500, 8.22)
            (1000, 8.45)
            };
            \addlegendentry{Token-level}
        \end{axis}
    \end{tikzpicture}
    \end{minipage}}
    }
    \subfigure[U-CER]{
    \label{fig:effect_keyword_list_ucer}
    \scalebox{0.5}{
    \begin{minipage}{0.45\textwidth}
    \centering
    \begin{tikzpicture}
        \centering
        \begin{axis}[
            xmin=0,xmax=1000,
            ymin=3.5,ymax=5.0,
            xtick={0, 50, 100, 200, 500, 1000},
            x tick label style={rotate=45,anchor=east},
            ytick={3.6, 3.8, 4, 4.2, 4.4, 4.6, 4.8, 5.0},
            ymajorgrids=true,
            major grid style={line width=.2pt,draw=gray!50,densely dashed},
            minor grid style={line width=.1pt,line length=.02pt,draw=gray!30},
            legend style={at={(0.75,0.5)}, anchor=south},
            ]
            \addplot[color=myorange, line width=1.5pt, mark=diamond, smooth] coordinates {
            (0, 3.94)
            (50, 3.82)
            (100, 3.89)
            (200, 3.86)
            (500, 4.03)
            (1000, 4.01)
            };
            \addlegendentry{Joint}
            \addplot[color=mygreen, line width=1.5pt, mark=triangle, smooth] coordinates {
            (0, 4.67)
            (50, 4.68)
            (100, 4.71)
            (200, 4.73)
            (500, 4.85)
            (1000, 4.84)
            };
            \addlegendentry{Phrase-level}
            \addplot[color=myred, line width=1.5pt, mark=square, smooth] coordinates {
            (0, 4.24)
            (50, 4.02)
            (100, 4.01)
            (200, 4.07)
            (500, 4.06)
            (1000, 4.18)
            };
            \addlegendentry{Token-level}
        \end{axis}
    \end{tikzpicture}
    \end{minipage}}
    }
    \subfigure[R]{
    \label{fig:effect_keyword_list_r}
    \scalebox{0.5}{
    \begin{minipage}{0.45\textwidth}
    \centering
    \begin{tikzpicture}
        \centering
        \begin{axis}[
            xmin=0,xmax=1000,
            ymin=84,ymax=97,
            xtick={0, 50, 100, 200, 500, 1000},
            x tick label style={rotate=45,anchor=east},
            ytick={84, 86, 88, 90, 92, 94, 96, 98},
            ymajorgrids=true,
            major grid style={line width=.2pt,draw=gray!50,densely dashed},
            minor grid style={line width=.1pt,line length=.02pt,draw=gray!30},
            legend style={at={(0.75,0.4)}, anchor=south},
            ]
            \addplot[color=myorange, line width=1.5pt, mark=diamond, smooth] coordinates {
            (0, 96.58)
            (50, 96.37)
            (100, 95.81)
            (200, 95.31)
            (500, 94.61)
            (1000, 94.02)
            };
            \addlegendentry{Joint}
            \addplot[color=mygreen, line width=1.5pt, mark=triangle, smooth] coordinates {
            (0, 95.46)
            (50, 95.46)
            (100, 95.02)
            (200, 94.61)
            (500, 93.84)
            (1000, 93.16)
            };
            \addlegendentry{Phrase-level}
            \addplot[color=myred, line width=1.5pt, mark=square, smooth] coordinates {
            (0, 91.60)
            (50, 86.94)
            (100, 86.09)
            (200, 85.56)
            (500, 84.88)
            (1000, 84.53)
            };
            \addlegendentry{Token-level}
        \end{axis}
    \end{tikzpicture}
    \end{minipage}}
    }
    \subfigure[RTF]{
    \label{fig:effect_keyword_list_rtf}
    \scalebox{0.5}{
    \begin{minipage}{0.45\textwidth}
    \centering
    \begin{tikzpicture}
        \centering
        \begin{axis}[
            ybar,
            bar width=8pt,
            xmin=0,xmax=3.5,
            ymin=0,ymax=6.5,
            xtick={0, 0.85,1.85,2.85},
            xticklabels={0, 50, 200, 1000},
            ytick={0, 1, 2, 3, 4, 5, 6},
            ymajorgrids=true,
            major grid style={line width=.2pt,draw=gray!50,densely dashed},
            minor grid style={line width=.1pt,line length=.02pt,draw=gray!30},
            legend style={at={(0.25,0.95)}, anchor=north},
            ylabel={RTF ($\times 10^{-2}$)},
            ylabel style={at={(axis description cs:+0.14,.5)}, anchor=south},
            ]
            \addplot[fill=myorange, draw=orange!70!black] coordinates {
                (0.85, 1.72)
                (1.85, 2.10)
                (2.85, 6.03)
            };
            \addplot[fill=mygreen, draw=green!60!black] coordinates {
                (0.85, 1.68)
                (1.85, 2.08)
                (2.85, 6.01)
            };
            \addplot[fill=myred, draw=red!60!black] coordinates {
                (0.85, 1.65)
                (1.85, 1.99)
                (2.85, 5.90)
            };

            \legend{Joint, Phrase-level, Token-level}
        \end{axis}
        \begin{axis}[
            xmin=0,xmax=3.5,
            ymin=0,ymax=6.5,
            xtick={0.85,1.85,2.85},
            xticklabels={},
            ytick={},
            axis x line=none,
            axis y line=none,
            legend style={at={(0.25,0.85)}, anchor=north},
            ]
            \addplot[color=gray, line width=1.2pt, mark=*, mark options={fill=gray}, smooth] coordinates {
                (0, 1.48)
                (0.85, 1.68)
                (1.85, 2.06)
                (2.85, 5.98)
            };
        \end{axis}
    \end{tikzpicture}
    \end{minipage}}
    }
    \caption{Effect of the keyword list size. The x-axis represents the number of keywords included in the keyword list. In Figures~\ref{fig:effect_keyword_list_bcer}, \ref{fig:effect_keyword_list_ucer}, and \ref{fig:effect_keyword_list_r}, the y-axes indicate B-CER, U-CER, and Recall, respectively, measured in percentage (\%). In Figure~\ref{fig:effect_keyword_list_rtf}, the y-axis shows the Real-Time Factor (RTF), measured in units of $\times 10^{-2}$. RTF is defined as the processing time divided by the duration of the input audio signal. The larger the RTF, the slower the system.}
    \label{fig:effect_keywordlist_new}
\end{figure*}
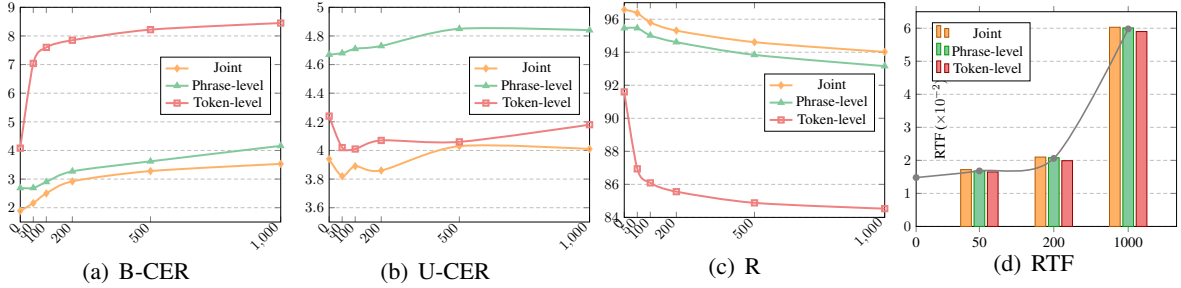

\subsubsection{Results on Chinese.}
First, we list the results of the fine-tuned Whisper, CopyNE, and our joint multi-grained model on the three Chinese datasets in Table \ref{table:res_zh}.
In the early experiments, we found that the original Whisper model performed poorly on the Chinese datasets, with a CER of approximately 14\% on the Aishell test set. 
Even after applying the text normalizer, the performance remained unsatisfactory\footnote{Similar results have been reported in some open-source repositories, such as \url{https://github.com/yeyupiaoling/Whisper-Finetune}.}. 
Therefore, the table does not include the original Whisper results and instead includes the fine-tuned version. The fine-tuned Whisper model achieved strong performance, making it a robust baseline for experimental comparison.

We evaluate the models on the test sets of Aishell, DC, and ICI. Since the models are fine-tuned on the Aishell training set, their performance on DC and ICI reflects their ability to generalize to out-of-domain scenarios. Table \ref{table:res_zh} presents the results across all three datasets.

Overall, our multi-grained model achieves the best performance across all metrics, particularly those related to keywords. Specifically, compared to CopyNE, our joint model achieves absolute reductions in B-CER of 1.2\%, 3.5\%, and 5.9\% on Aishell, DC, and ICI, respectively. Similarly, for Recall, our model outperforms CopyNE with absolute improvements of 2\%, 4.6\%, and 9.8\% on the three datasets.
Notably, the advantages of our method become even more pronounced when the keyword density is higher. For instance, in ICI, where the ratio of keyword characters reaches 18\%, the improvement achieved by our method is the largest among the three datasets.
While demonstrating outstanding performance on keyword-related metrics, our model also shows consistent improvements on general text, highlighting the robustness of our approach.

\subsubsection{Results on English.}
Table \ref{table:res_eng} shows the results on the test set of the English Slidespeech dataset. 
First, as shown in the first row of the table, the original Whisper model already achieves strong performance. 
Compared to the raw Whisper baseline, the fine-tuned version exhibits degraded performance on keyword-related metrics but achieves better WER and U-WER scores. We attribute this to the fact that the original Whisper, having been pre-trained on extensive English data, already possesses robust transcription capabilities. 
Further fine-tuning primarily adapts the model to the linguistic style of the Slidespeech dataset, which appears to come at the cost of its original keyword transcription accuracy.

Subsequently, by comparing with methods that leverage textual keywords, we observe that CPP and LCB-Net significantly lag behind CopyNE, MaLa-ASR, and our proposed multi-grained model.
CopyNE is a phrase-level fusion model whose backbone is also based on Whisper.
MaLa-ASR \footnote{For fair comparison, we evaluate MaLa-ASR's version without LLM fine-tuning.} is a token-level fusion model with 7B Vicuna \cite{chiang-2023-vicuna} as the LLM backbone.
Our model demonstrates superior performance on keyword-centric metrics compared to both baselines, while achieving comparable results to MaLa-ASR on non-keyword text segments. Notably, our approach maintains this competitive performance with greater efficiency, as it employs a compact 3.8B-parameter model compared to MaLa-ASR’s 7B LLM.

\subsubsection{Effect of Different Fusion Modules.}

To assess the contributions of token-level and phrase-level fusion in our multi-grained model, we perform ablation studies on the three Chinese datasets. The lower half of Table~\ref{table:res_zh} reports results after removing the phrase-level module (w/o P) or the token-level module (w/o T).

Removing phrase-level fusion leads to a notable drop in keyword-related metrics with little effect on non-keyword performance. In contrast, removing token-level fusion mainly degrades non-keyword accuracy, with limited impact on keyword recognition. Both variants still outperform the Whisper baseline, which lacks contextual biasing.

These results show that the two fusion strategies are complementary: phrase-level fusion boosts keyword recognition, while token-level fusion improves general transcription. The full multi-grained model combines both strengths for superior overall performance.

\subsubsection{Effect of the Keyword List Size.}
The size of the keyword list has a notable impact on model performance \cite{pundak-2018-clas,zhou-2023-copyne}. In practice, keyword dictionaries are often large and noisy, which may hinder model effectiveness. We analyze model robustness on Aishell under varying list sizes, focusing on B-CER, U-CER, and Recall, since overall CER remains stable.

As shown in Figure~\ref{fig:effect_keywordlist_new}, the multi-grained model consistently achieves the best performance across all list sizes. While all methods degrade with more keywords, token-level fusion shows the sharpest decline, indicating lower stability. Phrase-level fusion is more robust, and the multi-grained model inherits this advantage.

Interestingly, as shown in Figure~\ref{fig:effect_keyword_list_ucer}, the U-CER of the token-level model briefly improves when the list grows from 0 to 100. This may be due to richer prompts enhancing the LLM’s language modeling for non-keyword text. However, performance degrades as noise increases with larger lists.

Figure~\ref{fig:effect_keyword_list_rtf} shows the Real-Time Factor (RTF) across list sizes. Token-level fusion is the most efficient; the multi-grained model is slightly slower due to its more complex design, but the trade-off is acceptable given its performance. While RTF rises significantly at 1000 keywords, it remains below 0.06, meeting real-time requirements. We also mark the Whisper baseline (RTF = 0.0148) at list size 0 in the figure; notably, our method with a list size of 50 yields a comparable RTF, indicating that incorporating a moderate number of keywords introduces minimal additional latency. In practice, keyword prompts can be reused across queries, further reducing runtime overhead.

\paragraph{Impact of LLM Size.}
\begin{table}[!tb]
    \centering
    \setlength{\tabcolsep}{2pt}
    \begin{tabular}{l cccc}
        \toprule
        Model & CER$\downarrow$ & B-CER$\downarrow$ & U-CER$\downarrow$ & R$\uparrow$ \\
        \midrule
        Qwen2-1.5B & 3.7 & 2.2 & 3.8 & 96.4 \\
        Qwen2-7B & 3.6 & 2.2 & 3.7 & 96.4 \\
        \bottomrule
    \end{tabular}
    \caption{Results of LLM models with different sizes on the Aishell dataset.} 
    \label{table:2bvs7b}
\end{table}
Larger LLMs generally offer better performance, so we explored how different LLM sizes affect our multi-grained fusion model. Table~\ref{table:2bvs7b} compares the results on the Aishell dataset using Qwen2-1.5B and Qwen2-7B as the backbone.
Actually, we did not observe notable performance gains from the larger 7B model. It achieved only a marginal 0.1\% improvement in U-CER and CER compared to the 1.5B model. 
This suggests that in the Contextual ASR task, where the primary challenge lies in integrating keyword prompts into transcription, a smaller LLM like Qwen2-1.5B already possesses sufficient capacity to capture and leverage the necessary contextual information. The minimal improvement from scaling up to 7B indicates diminishing returns with respect to LLM size in this specific setting.
Considering the trade-off between performance and efficiency, the 1.5B model is a more practical choice.

\section{Conclusion}

In this paper, we propose a multi-grained fusion approach that combines token-level and phrase-level fusion to enhance keyword recognition in ASR systems. 
Through the introduction of a late-fusion mechanism, our approach effectively leverages the advanced contextual modeling capabilities of LLMs to assist ASR models, achieving superior accuracy on keywords while preserving robust performance on non-keyword text.
We believe our work paves the way for novel applications through the fusion of LLMs and ASR, providing a strong foundation for future research in ASR keyword recognition.

\bibliography{custom}

\end{CJK}
\end{document}